\documentclass{article}

\usepackage[preprint]{neurips_2026}

\usepackage[utf8]{inputenc} 
\usepackage[T1]{fontenc}    
\usepackage{hyperref}       
\usepackage{url}            
\usepackage{booktabs}       
\usepackage{amsfonts}       
\usepackage{nicefrac}       
\usepackage{microtype}      
\usepackage{xcolor}         
\usepackage{graphicx}
\usepackage{multirow}
\usepackage{amsmath}
\usepackage{tabularx}

\newcommand{\calvin}{CALVIN ABC\raisebox{0.15ex}{$\rightarrow$}D}
\title{Mitigating State Aliasing in Vision-Language-Action Models via Inverse Dynamics Learning}

\author{%
  Kyujin Lee\\
  KAIST\\
  \texttt{kyujinlee@kaist.ac.kr} \\
  \And
  Injae Kim\\
  KAIST\\
  \texttt{injae.kim@kaist.ac.kr} \\
  \And
  Jihwan Park\\
  KAIST\\
  \texttt{jseven7071@kaist.ac.kr} \\
  \And
  Yejun Ju\\
  KAIST\\
  \texttt{dpwns99@kaist.ac.kr} \\
  \And
  Minseok Joo\\
  Korea University\\
  \texttt{wlgkcjf87@korea.ac.kr} \\
  \And
  Hyunwoo J. Kim\thanks{Corresponding author.} \\
  KAIST\\
  \texttt{hyunwoojkim@kaist.ac.kr} \\
}

\begin{document}

\maketitle

\begin{abstract}
Vision-Language-Action (VLA) models have emerged as a promising framework that unifies perception, reasoning, and control for handling robot manipulation tasks by adapting pretrained vision-language models (VLMs) to action prediction. 
However, VLM-derived representations are often insensitive to subtle visual distinctions required for low-level control, causing \textit{state aliasing} between visually similar states but substantially different in required actions.
Prior VLA studies improve visual understanding by training models to generate visual or reasoning outputs, such as future frames, 2D grounding points or traces, or intermediate spatial reasoning steps, but these objectives typically shape the vision encoder only indirectly through end-to-end output prediction, and state aliasing is not explicitly analyzed in the learned visual feature space.
To mitigate state aliasing, we introduce \textbf{inverse dynamics learning} as an auxiliary objective that directly supervises the VLA vision encoder.
By predicting the action between current and future observations, our objective encourages the encoder to capture fine-grained visual distinctions that determine the required low-level actions.
We further use pseudo-reversed supervision to expose the encoder to a broader range of action directions and improve generalization under limited robot demonstrations.
Our method is readily applicable to diverse VLA baselines, as it uses only observation-action pairs from standard manipulation datasets without additional annotations during training and preserves the original inference pipeline at test time.
Experiments on CALVIN ABC\raisebox{0.15ex}{$\rightarrow$}D and SimplerEnv demonstrate consistent performance gains across diverse VLA baselines.
We further demonstrate, through frozen-encoder probing and state-feature alignment analyses, that our method learns state-discriminative visual representations that reduce state aliasing and better align with robot state changes.
\end{abstract}

\section{Introduction}
\begin{figure}[t!]
        \centering
        \includegraphics[width=1\linewidth]{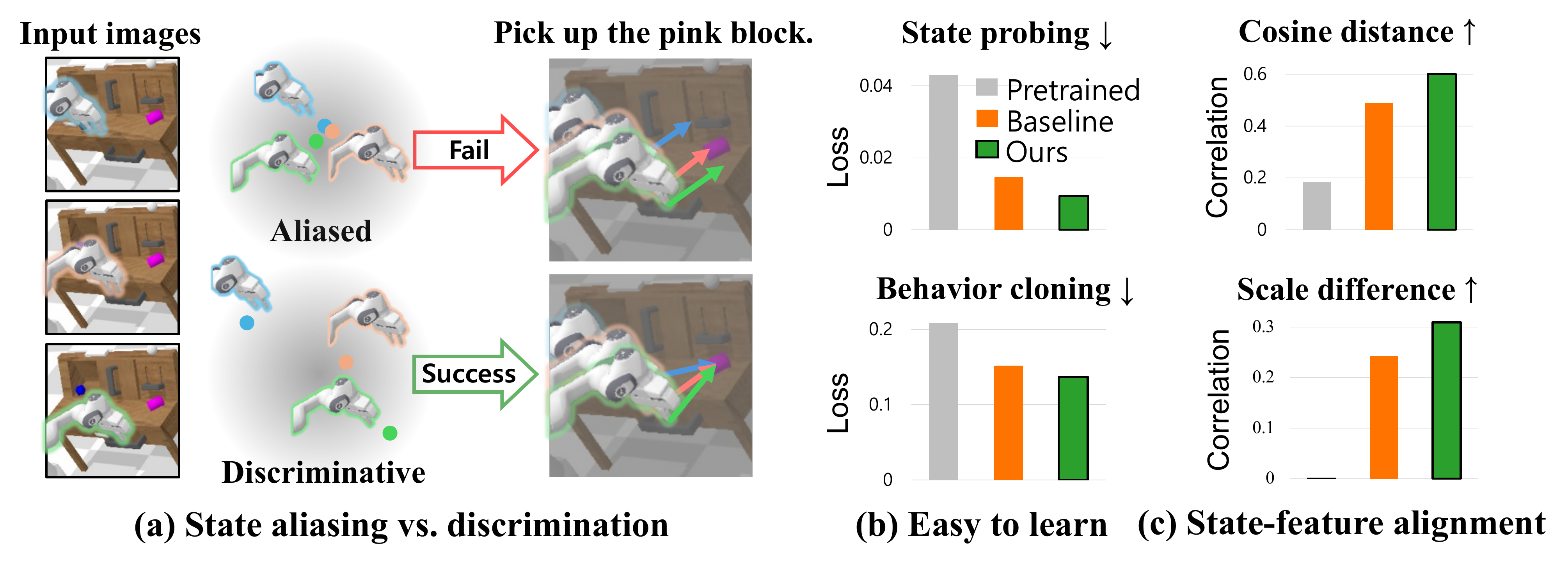}
        \caption{\textbf{Motivation and representation analysis.}
(a) State aliasing maps visually similar states requiring different actions to nearby representations, causing ambiguous action prediction; state-discriminative representations instead separate relevant states and preserve meaningful robot-state structure.
(b) Lower frozen-probe training loss indicates easier behavior and state prediction from the learned features, consistent with reduced state aliasing.
(c) Higher structured state alignment suggests that our visual representations distinguish robot states while preserving their underlying state-space structure. 
Details are provided in Secs.~\ref{exp:ve_evaluation} and~\ref{sec4.5:analysis}.}
        \vspace{-10pt}
        \label{fig1:correlation}
\end{figure}

Vision-Language-Action (VLA) models have emerged as a promising framework for robot manipulation, as they map visual observations and language instructions to executable robot actions while leveraging the semantic priors of pretrained vision-language models (VLMs) \citep{liu2023visual, karamcheti2024prismatic, beyer2024paligemma, bai2025qwen3}.
Despite this promise, robotic manipulation requires visual representations that go beyond high-level vision-language understanding.
A policy must distinguish fine-grained state variations, such as the end-effector pose, object pose, and their spatial relation, which may be visually subtle but determine the correct low-level action.
Because VLM-derived representations are initially optimized for large-scale vision-language understanding, they primarily capture high-level semantic concepts and are insensitive to fine-grained spatial distinctions required for low-level control \citep{zheng2024lm4lv, chen2024spatialvlm, wen2024can}.
This can lead to \textbf{state aliasing} Fig.~\ref{fig1:correlation} (a), where visually similar but distinct states are mapped to insufficiently discriminative features, causing ambiguous action prediction.
This motivates learning \textbf{state-discriminative visual representations} for reliable VLA policies.

Prior VLA studies have improved visual understanding for manipulation by training models to generate additional visual or reasoning targets, including visual prediction \citep{wu2023unleashing, zhang2025up, zhao2025cot, zhang2025dreamvla, wang2025unified}, 2D grounding \citep{niu2024llarva, yang2025magma, lee2025molmoact, chen2026vista, yuan2025embodied, wang2025trackvla}, and spatial reasoning before action prediction \citep{zawalski2024robotic, sun2025emma, huang2025thinkact}.
These decoding-oriented objectives provide useful supervision for the overall policy, but they are typically optimized end-to-end through output generation, leaving the vision encoder only indirectly supervised.
Consequently, state aliasing remains underexplored: the encoder is not primarily trained or analyzed to distinguish visually similar states that require different actions.

To mitigate state aliasing in VLA policies, we introduce \textbf{inverse dynamics learning} as an auxiliary objective that directly supervises the vision encoder during training.
Given current and future observations from a demonstration trajectory, the vision encoder independently extracts their visual features, and a lightweight inverse dynamics head predicts the intermediate action.
This supervision encourages the encoder to retain action-grounded visual distinctions that determine low-level actions, making visually similar states that require different actions more separable in the feature space.
As shown in Fig.~\ref{fig1:correlation} (b), compared with standard VLA training, our objective yields visual representations from which downstream policies can learn more easily, suggesting reduced state aliasing.
Moreover, in Fig.~\ref{fig1:correlation} (c), the feature space is not merely more separable, but also better aligned with actual robot state changes.
These representation gains are obtained with an auxiliary head that is used only during training and removed afterward, leaving the original VLA architecture and inference process intact.

To further strengthen the inverse dynamics learning, we introduce \textbf{Pseudo Time Reversal} (PTR), an augmentation for inverse dynamics learning that addresses the one-directional nature of robot demonstrations.
Because robot datasets typically contain only actions along successful trajectories, the model sees limited action variation around similar visual states.
This can encourage the vision encoder to rely on coarse directional shortcuts, such as trajectory direction or task progress, rather than action-relevant fine-grained visual distinctions, weakening generalization beyond the demonstrated action patterns \citep{park2022robust, shao2024offline, xing2025shortcut}.
PTR constructs pseudo-reversed supervision from the same demonstrations by temporally reversing the observation order and pairing it with pseudo-ground-truth reversed actions. This exposes the encoder to both forward and pseudo-reversed action directions, expanding action diversity and reducing reliance on directional shortcuts in vision encoder learning.

Extensive experiments on \calvin~\citep{mees2022calvin} and SimplerEnv~\citep{li2024evaluating} show that our method consistently improves end-to-end policy performance across diverse VLA baselines.
Frozen-encoder evaluations on LIBERO~\citep{liu2023libero} further show that the learned representations support easier downstream action and robot state prediction, while state-feature alignment analysis demonstrates stronger correspondence between visual feature distances and ground-truth robot state changes.
Together, these results indicate that our method mitigates state aliasing by learning visual features that are both more separable and more aligned with robot state geometry. 

In summary, our contributions are threefold:

\begin{itemize}
    \item We introduce a simple yet effective pretext task, inverse dynamics learning, that predicts actions between visual observations, encouraging a vision encoder in VLA to learn action-aware visual representations that mitigate state aliasing.
    \item We propose Pseudo Time Reversal (PTR), a data augmentation method that temporally reverses visual observations along with pseudo-ground-truth actions for more robust generalization under limited action diversity in low-demonstration regimes.
    \item Our experiments demonstrate that the proposed method provides consistent performance gains across diverse VLA baselines. The effectiveness of learned representations is also validated through frozen-encoder probing and state-feature alignment analysis.
\end{itemize}

\section{Related Work}
\subsection{Vision-language-action models}

Vision-language-action (VLA) models adapt pretrained vision-language models (VLMs) \citep{liu2023visual, karamcheti2024prismatic, beyer2024paligemma, bai2025qwen3} to map visual observations and language instructions to low-level robot actions, establishing VLM initialization as a central design choice for generalist robot policies \citep{zitkovich2023rt, black2024pi_0, kim2024openvla, yang2025magma}. 
Recent studies have explored diverse policy formulations, including autoregressive token prediction \citep{zitkovich2023rt, kim2024openvla, wang2025unified}, continuous action prediction via direct regression, diffusion, or flow matching \citep{black2024pi_0, li2024cogact, wen2024diffusion}, spatially grounded action prediction \citep{qu2025spatialvla, li2025bridgevla}, and hybrid discrete-continuous action modeling \citep{liu2025hybridvla, driess2025knowledge}. 
Another line of work improves VLM-to-VLA adaptation through fine-tuning strategies, lightweight adapters, and compact policy architectures \citep{liu2025towards, kim2025fine, reuss2025flower, wang2026vla}.

\subsection{Visual and reasoning supervision in VLA}

Beyond policy architectures, recent work has explored supervision for visual understanding in VLA models. One line augments action learning with visual prediction objectives, such as future-frame or depth prediction, to provide spatial and temporal supervision beyond action prediction \citep{wu2023unleashing, zitkovich2023rt, zhao2025cot, zhang2025up, zhang2025dreamvla}. Another introduces grounding and reasoning objectives that anchor action prediction to fine-grained visual evidence or intermediate plans, including pixel- and trace-level supervision over task-relevant regions \citep{niu2024llarva, yuan2025embodied, yang2025magma} and chain-of-thought, latent, or grounded reasoning prior to action \citep{zawalski2024robotic, huang2025thinkact, yang2025instructvla, sun2025emma, intelligence2025pi_, huang2025motvla, huang2026fast}. 
Robot-state-aware contrastive learning has also been used to align VLA representations with proprioceptive states \citep{kim2025contrastive}. 
Although these approaches provide useful supervision, they are typically optimized through generated visual or reasoning outputs, leaving the vision encoder only indirectly supervised. In contrast, our method applies inverse dynamics supervision to the vision encoder itself to address state aliasing in the learned visual features.

\subsection{Inverse dynamics modeling}

Inverse dynamics predicts the action between two observations and has been used in several learning-based robotics settings. 
In predictive policy learning, it maps predicted future visual states to executable actions \citep{du2023learning, tian2024predictive, hu2024video, zhang2026disentangled}. 
Another line of work uses inverse dynamics to recover actions or latent actions from action-free videos, enabling policy learning without explicit action labels \citep{schmidt2023learning, ye2024latent, yang2025learning, routray2025vipra}. 
Closest to our work, inverse dynamics has been studied for imitation learning and visual representation learning, where predicting actions from state transitions helps preserve control-relevant information \citep{brandfonbrener2023inverse, li2024robust, kim2024investigating, cui2024dynamo}. 
We adapt this idea for VLA vision encoder learning and introduce PTR to regularize representations under limited vision-action diversity.

\section{Method}
\begin{figure}[t!]
        \centering
        \includegraphics[width=1\linewidth]{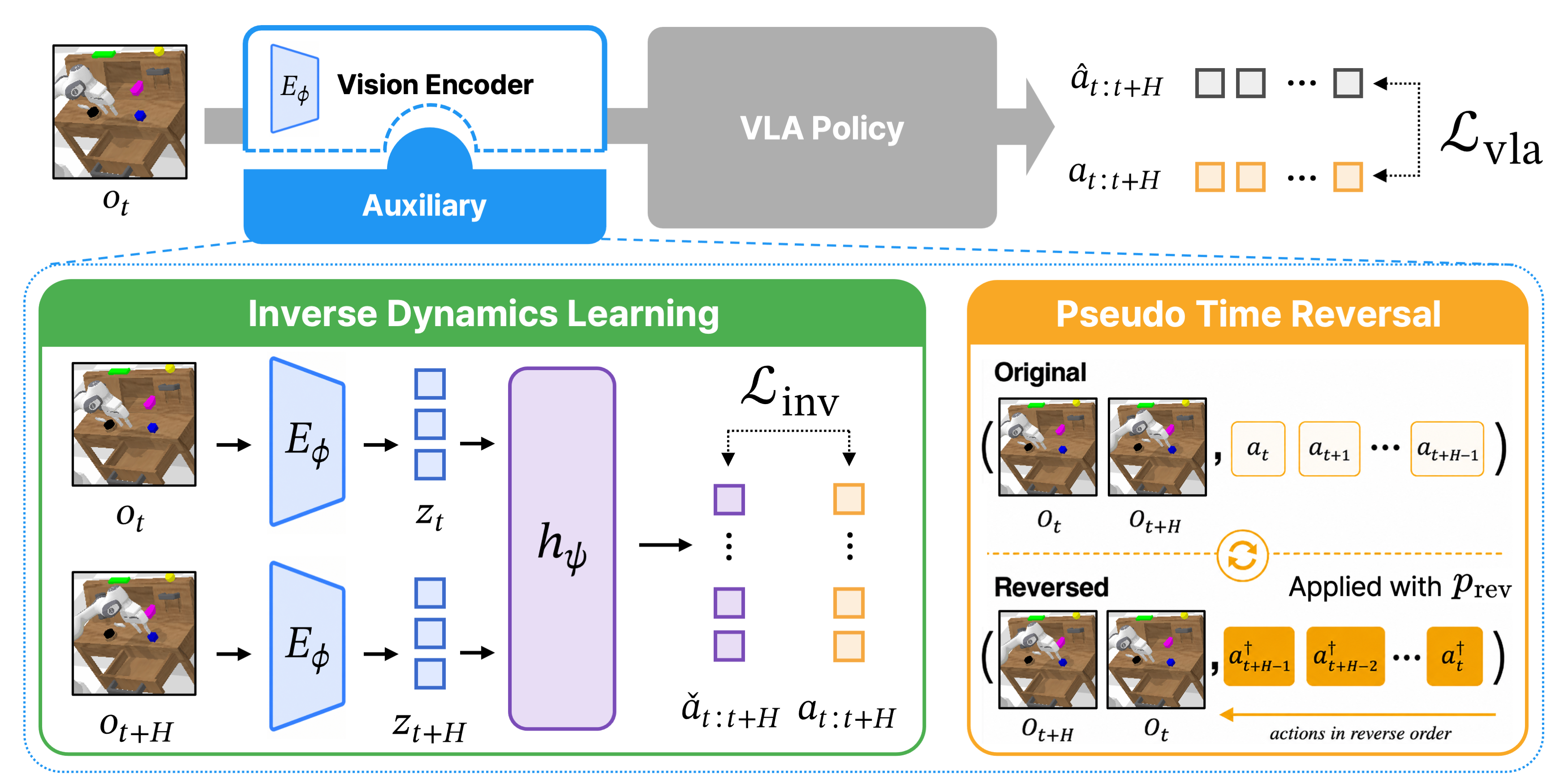}
        \caption{\textbf{Method overview.}
To mitigate state aliasing in VLA policies, we supervise the vision encoder with inverse dynamics learning during training.
In addition to the main VLA objective, the vision encoder \(E_\phi\) independently encodes current and future observations \((o_t,o_{t+H})\), and the inverse dynamics head \(h_\psi\) predicts the intermediate action sequence.
This auxiliary supervision encourages the encoder to retain visual distinctions that determine low-level actions.
PTR further constructs pseudo-reversed supervision by swapping the observation order and reversing the action sequence with negated motion offsets, improving generalization under limited action diversity.
After training, the auxiliary head is removed, leaving the original VLA architecture and inference process intact.}
\end{figure}
We propose inverse dynamics learning as an auxiliary objective that directly supervises the vision encoder to mitigate state aliasing. 
Our method encourages state-discriminative visual representation learning while keeping the main policy intact. 
We further introduce Pseudo Time Reversal (PTR) to improve generalization under limited action diversity in robot demonstrations. 
Sec.~\ref{met:aux_invdyn} presents the inverse dynamics objective, Sec.~\ref{met:ptr} introduces PTR, and Sec.~\ref{met:pipeline} describes their integration with the main VLA objective.

\subsection{Auxiliary inverse dynamics learning}
\label{met:aux_invdyn}

Inverse dynamics is the task of predicting the action between two observations.
We use this task as an auxiliary supervision signal for the VLA vision encoder, encouraging it to preserve visual differences that determine low-level robot actions.

We consider a language-conditioned robot manipulation dataset $\mathcal{D} = \{\tau_i\}_{i=1}^{N}$ consisting of demonstration trajectories. For a trajectory $\tau \in \mathcal{D}$, we write
$$\tau = \left(l,\{(o_t, a_t)\}_{t=1}^{T}\right),$$
where $l$ denotes a language instruction, $o_t$ denotes the visual observation at time step $t$, and $a_t$ denotes the corresponding low-level robot action. Here, $o_t$ may consist of one or more camera views. While the action space can vary across embodiments, we consider a 7-DoF single-gripper action space in this work:
$$a_t = [\Delta x_t, \Delta y_t, \Delta z_t, \Delta \phi_t, \Delta \theta_t, \Delta \psi_t, g_t],$$
where $\Delta x_t, \Delta y_t, \Delta z_t$ denote the relative translation offsets of the end-effector, $\Delta \phi_t, \Delta \theta_t, \Delta \psi_t$ denote the relative rotation offsets, and $g_t$ denotes the gripper open/close action.

For an action chunk of length $k$, we use the notation
\[
a_{t:t+k} = (a_t,a_{t+1},\dots,a_{t+k-1}).
\]

Given an observation pair $(o_t,o_{t+k})$ and the corresponding action chunk $a_{t:t+k}$, the vision encoder $E_{\phi}$ independently encodes the two observations into visual features $z_t$ and $z_{t+k}$:
\[
z_t = E_{\phi}(o_t), \qquad z_{t+k} = E_{\phi}(o_{t+k}).
\]
A lightweight inverse dynamics head $h_{\psi}$ then predicts the intermediate action chunk:
\[
\check{a}_{t:t+k} = h_{\psi}(z_t,z_{t+k}),
\]
where $\check{a}_{t:t+k}$ denotes the inverse dynamics prediction.

We supervise this prediction with an auxiliary inverse dynamics loss:
\[
\mathcal{L}_{\mathrm{inv}}
=
\mathcal{L}_{\mathrm{motion}}(\check{a}_{t:t+k},a_{t:t+k})
+
\lambda_g
\mathcal{L}_{\mathrm{gripper}}(\check{a}_{t:t+k},a_{t:t+k}).
\]
Here, $\mathcal{L}_{\mathrm{motion}}$ is an $\ell_1$ loss applied to the six end-effector motion offsets, $\mathcal{L}_{\mathrm{gripper}}$ is a binary cross-entropy loss applied to the gripper action, and $\lambda_g$ balances the two terms.

Since the auxiliary pairs are constructed directly from observation-action sequences already available in standard robot manipulation datasets, inverse dynamics learning requires no additional annotations while providing supervision that helps mitigate state aliasing in the vision encoder.

\subsection{Pseudo Time Reversal}
\label{met:ptr}

Robot manipulation datasets typically provide supervision only along successful forward trajectories. As a result, for a given image-language condition, the available actions often cover limited directional variation, which can encourage the vision encoder to rely on coarse directional patterns or task-irrelevant visual shortcuts rather than manipulation-relevant cues \citep{xing2025shortcut}. To mitigate this bias, we introduce Pseudo Time Reversal (PTR), a simple stochastic augmentation for auxiliary inverse dynamics learning.

Given an inverse dynamics sample $(o_t, o_{t+k}, a_{t:t+k})$, PTR constructs a pseudo-reversed sample by swapping the observation order and reversing the action sequence after negating its motion offsets. For each action $a_i$, we define
\[
a_i^{\dagger}
=
[-\Delta x_i, -\Delta y_i, -\Delta z_i, -\Delta \phi_i, -\Delta \theta_i, -\Delta \psi_i, g_i].
\]
Correspondingly, the pseudo-reversed counterpart of the action chunk $a_{t:t+k}$ is defined as
\[
a_{t+k:t}^{\dagger}
=
(a_{t+k-1}^{\dagger}, a_{t+k-2}^{\dagger}, \dots, a_t^{\dagger}).
\]

PTR is applied to each auxiliary sample with probability $p_{\mathrm{rev}}$. The sample is kept as $(o_t, o_{t+k}, a_{t:t+k})$ with probability $1-p_{\mathrm{rev}}$, and converted to $(o_{t+k},o_t,a_{t+k:t}^{\dagger})$ with probability $p_{\mathrm{rev}}$. This exposes the inverse dynamics objective to both forward and pseudo-reversed supervision.

The reversal is pseudo because negating the motion offsets produces motion in the opposite direction with respect to the original observation, rather than the physically exact action that returns from $o_{t+H}$ to $o_t$. 
Since PTR is used only as an auxiliary signal, not as a target for the main VLA policy, this local approximation is sufficient to regularize visual representation learning without requiring reversed demonstrations.

\subsection{Overall pipeline}
\label{met:pipeline}

We now describe how auxiliary inverse dynamics learning and PTR are integrated into standard VLA training. 
Let $\pi_{\theta}$ denote a VLA policy that predicts an action chunk $\hat{a}_{t:t+H}$ of length $H$ from the current observation $o_t$ and language instruction $l$:
\[
\hat{a}_{t:t+H} = \pi_{\theta}(o_t,l).
\]
The policy is trained with its original action prediction objective:
\[
\mathcal{L}_{\mathrm{vla}}
=
\ell(\hat{a}_{t:t+H},a_{t:t+H}),
\]
where $\ell$ denotes the model-specific action prediction loss.

For the auxiliary inverse dynamics objective, we use the same action chunk as the VLA target by setting $k=H$. 
Thus, each VLA training sample $(l,o_t,a_{t:t+H})$ is augmented with the future observation $o_{t+H}$ from the same trajectory to form an inverse dynamics sample $(o_t,o_{t+H},a_{t:t+H})$. 
This auxiliary sample is then stochastically kept in its forward form or converted into its pseudo-reversed form with probability $p_{\mathrm{rev}}$, as described in Sec.~\ref{met:ptr}.

Using either the forward sample or its pseudo-reversed variant, we compute the auxiliary inverse dynamics loss $\mathcal{L}_{\mathrm{inv}}$ defined in Sec.~\ref{met:aux_invdyn}. The overall training objective is
\[
\mathcal{L}
=
\mathcal{L}_{\mathrm{vla}}
+
\lambda_{\mathrm{inv}}\mathcal{L}_{\mathrm{inv}},
\]
where $\lambda_{\mathrm{inv}}$ controls the strength of the auxiliary supervision.

Importantly, our method does not modify the policy architecture or its action prediction objective. 
The inverse dynamics head is attached only during training to provide additional supervision to the vision encoder, and is removed after training. 
At inference time, the original VLA policy $\pi_{\theta}$ is used intact, taking only the current observation and language instruction as input.

\section{Experiments}
We evaluate our method from two perspectives: end-to-end policy performance across various VLA architectures and whether the learned vision encoder reduces state aliasing in its visual representations. For policy evaluation, we benchmark our method on \calvin~\citep{mees2022calvin} and SimplerEnv (WidowX) \citep{li2024evaluating} using multiple VLA baselines with different modeling choices. For representation analysis, we conduct frozen-encoder evaluations on LIBERO \citep{liu2023libero}, comparing the original Qwen3VL-8B \citep{bai2025qwen3} vision encoder, VLM4VLA \citep{zhang2026vlm4vla} after LIBERO-90 pretraining, and VLM4VLA + Ours. These evaluations test whether the learned visual features make downstream action and robot state prediction easier, which serves as evidence of reduced state aliasing. We further analyze state-feature alignment to examine whether visual feature distances better reflect robot-state changes, rather than merely separating observations. Finally, we include an ablation study to analyze the contribution of PTR.

\subsection{Experimental setup}

\textbf{Baseline models.} We compare our method against representative VLA baselines with different architectures and action modeling strategies. \textbf{VLM4VLA} \citep{zhang2026vlm4vla} is a simple VLA framework built from a pretrained VLM with minimal architectural modification, predicting continuous actions via direct regression. \textbf{FLOWER} \citep{reuss2025flower} is an efficient flow-based VLA model with a compact architecture and competitive action modeling performance. \textbf{SpatialVLA} \citep{qu2025spatialvla} predicts actions through explicit 3D spatial modeling, allowing us to evaluate our method under a different inference structure.

For each benchmark, we compare each baseline with its +Ours variant: VLM4VLA and FLOWER on \calvin, and SpatialVLA on SimplerEnv. For LIBERO representation analysis, we compare Qwen3VL-8B, VLM4VLA, and VLM4VLA + Ours in the frozen-encoder setting.

\textbf{Training protocol.} 
For each baseline, we keep the original training pipeline unchanged and add only auxiliary inverse dynamics learning and PTR to supervise the vision encoder. 
To ensure a fair comparison, each baseline and its +Ours variant are trained under matched training budgets, hyperparameters, checkpoint selection rules, and evaluation protocols. 
We follow released configurations when available, use default implementation values for unspecified details, and report final-checkpoint performance. 
The auxiliary loss weight is tuned separately for each baseline. 
Detailed hyperparameters and benchmark-specific training budgets are provided in Appendix~\ref{app:implementation_details}.

\subsection{\calvin}
\label{exp:calvin}
\begin{table}[t]
  \centering
  \small
  \caption{Results on \calvin. We report success rates over 1000 rollouts for completing 1--5 tasks in a row and the average completed sequence length. Bold numbers indicate the best results in each column. $^\ast$ indicates results reproduced in our environment.}
  \label{tab:calvin_results}
  \resizebox{\linewidth}{!}{
  \begin{tabular}{lcccccc}
    \toprule
    \multirow{2}{*}{Method} 
    & \multicolumn{5}{c}{Tasks completed in a row} 
    & \multirow{2}{*}{Avg. Len. $\uparrow$} \\
    \cmidrule(lr){2-6}
    & 1 & 2 & 3 & 4 & 5 & \\
    \midrule
    GR-1 \citep{wu2023unleashing}      & 85.4 & 71.2 & 59.6 & 49.7 & 40.1 & 3.06 \\
    UP-VLA \citep{zhang2025up}         & 92.8 & 86.5 & 81.5 & 76.9 & 69.9 & 4.08 \\
    RoboVLMs \citep{liu2025towards}    & 98.0 & 93.6 & 85.4 & 77.8 & 70.4 & 4.25 \\
    Seer \citep{tian2024predictive}    & 96.3 & 91.6 & 86.1 & 80.3 & 74.0 & 4.28 \\
    VPP \citep{hu2024video}            & 96.5 & 90.9 & 86.6 & 82.0 & 76.9 & 4.33 \\
    VLA-Adapter \citep{wang2026vla}    & 99.1 & 94.6 & 88.8 & 82.8 & 76.5 & 4.42 \\
    DreamVLA \citep{zhang2025dreamvla} & 98.2 & 94.6 & 89.5 & 83.4 & 78.1 & 4.44 \\
    \midrule
    VLM4VLA$^\ast$ \citep{zhang2026vlm4vla}
               & 93.4 & 85.9 & 79.7 & 75.2 & 69.2 & 4.03 \\
    + Ours     & 94.4 {\scriptsize (+1.0)}
               & 86.7 {\scriptsize (+0.8)}
               & 81.5 {\scriptsize (+1.8)}
               & 76.6 {\scriptsize (+1.4)}
               & 70.5 {\scriptsize (+1.3)}
               & 4.10 {\scriptsize (+0.07)}\\
    \midrule
    FLOWER \citep{reuss2025flower}
               & 99.3 & 95.9 & 90.5 & 84.8 & 77.5 & 4.54 \\
    + Ours     & \textbf{99.5} {\scriptsize (+0.2)}
               & \textbf{96.6} {\scriptsize (+0.7)}
               & \textbf{91.2} {\scriptsize (+0.7)}
               & \textbf{86.9} {\scriptsize (+2.1)}
               & \textbf{81.6} {\scriptsize (+4.1)}
               & \textbf{4.56} {\scriptsize (+0.02)} \\
    \bottomrule
  \end{tabular}
  }
\end{table}

We evaluate our method on CALVIN ABC\raisebox{0.15ex}{$\rightarrow$}D, a language-conditioned long-horizon manipulation benchmark. Following the standard protocol, models are trained on the ABC split and evaluated in the unseen D environment over 1000 rollouts. We report success rates for completing 1--5 consecutive subtasks, where evaluation terminates after the first failure, together with the average success sequence length.

As shown in Table~\ref{tab:calvin_results}, our method improves both VLM4VLA and FLOWER on \calvin. These gains show that directly supervising the vision encoder with inverse dynamics improves long-horizon manipulation performance across different VLA architectures. Notably, the improvement on FLOWER indicates that even a strong action-modeling baseline can benefit from better visual representation learning, suggesting that vision encoder supervision addresses a remaining bottleneck beyond advances in action decoding.

\subsection{SimplerEnv-Bridge}
\label{exp:simpler}
\begin{table}[t]
  \centering
  \small
  \caption{Results on the SimplerEnv Bridge benchmark. We report task success rates over 24 trials for each object category and the average success rate across categories. Bold numbers indicate the best results in each column. $^\ast$ indicates results reproduced in our environment.}
  \label{tab:simplerenv_bridge}
  \begin{tabular}{lccccc}
    \toprule
    Method & Carrot & Eggplant & Spoon & Cube & Average $\uparrow$ \\
    \midrule
    OpenVLA \citep{kim2024openvla} & 0.0  & 0.0  & 0.0  & 4.1  & 1.0  \\
    RoboVLMs (pre-trained) \citep{liu2025towards}     & \textbf{25.0} &  0.0 & 20.8 & 8.3 & 13.5 \\
    RoboVLMs (fine-tuned) \citep{liu2025towards}     & \textbf{25.0} & 58.3 & 20.8 & 12.5 & 31.3 \\
    SpatialVLA (pre-trained) \citep{qu2025spatialvla} & 20.8 & 54.2 & 12.3 & \textbf{20.8} & 27.0 \\
    \midrule
    SpatialVLA$^\ast$ (fine-tuned) \citep{qu2025spatialvla} 
               & 12.5 & 66.6  & \textbf{25.0} & 16.7 & 30.2 \\
    + Ours     & 12.5  
               & \textbf{75.0} {\scriptsize (+8.4)} 
               & \textbf{25.0}  
               & \textbf{20.8} {\scriptsize (+4.1)} 
               & \textbf{33.3} {\scriptsize (+3.1)}  \\
    \bottomrule
  \end{tabular}
\end{table}

We next evaluate our method on SimplerEnv (WidowX) using SpatialVLA trained on Bridge. Following the SpatialVLA evaluation protocol, we evaluate on four manipulation tasks and report final task success over 24 trials.

As shown in Table~\ref{tab:simplerenv_bridge}, our method improves SpatialVLA on SimplerEnv. These results show that inverse dynamics supervision for the vision encoder remains effective when trained on real-world robot data. The gain on a spatially structured VLA model further suggests that improving visual representation learning remains beneficial beyond standard action prediction architectures.

\subsection{Frozen vision encoder evaluation}
\label{exp:ve_evaluation}

While Secs.~\ref{exp:calvin} and \ref{exp:simpler} show improved policy performance, they do not directly reveal whether the vision encoder itself learns better representations. To address this, we perform frozen-encoder evaluations on LIBERO, inspired by \cite{brandfonbrener2023inverse}, to probe how well the learned visual features support downstream behavior cloning and state prediction.

\textbf{Evaluation protocol.}
We compare three frozen vision encoders: the original Qwen3VL-8B encoder used to initialize VLM4VLA, VLM4VLA after LIBERO-90 pretraining, and VLM4VLA + Ours after the same pretraining. For each encoder, we train only lightweight MLP heads while keeping the vision encoder fixed. We evaluate the frozen features with three probes: vision-only behavior cloning success, behavior cloning train/validation loss, and proprioceptive state prediction. Detailed training configurations are provided in Appendix~\ref{app:implementation_details}.

\textbf{Behavior cloning with frozen visual features.}
\label{ex:bc_success}
\begin{table}[t]
  \centering
  \small
  \caption{Frozen-encoder behavior cloning evaluation on LIBERO. We compare the original Qwen3VL-8B vision encoder, VLM4VLA, and VLM4VLA + Ours after LIBERO-90 pretraining. We run 20 trials per task, corresponding to 200 trials per suite. Bold numbers indicate the best results in each column.}
  \label{tab:libero_results}
  \begin{tabular}{lccccc}
    \toprule
    Vision Encoder & Spatial & Object & Goal & Long & Average $\uparrow$ \\
    \midrule
    Qwen3VL-8B \citep{bai2025qwen3} & 49.5\% & 39.5\% & 43.0\% & 9.5\% & 35.4\% \\
    VLM4VLA \citep{zhang2026vlm4vla}& \textbf{57.0\%} & 39.5\% & 55.0\% & 18.0\% & 42.4\% \\
    + Ours  & \textbf{57.0\%} & \textbf{43.5\%} {\scriptsize (+4.0)}  & \textbf{63.0\%} {\scriptsize (+8.0)}  & \textbf{19.5\%} {\scriptsize (+1.5)}& \textbf{45.8\%} {\scriptsize (+3.4)} \\
    \bottomrule
  \end{tabular}
\end{table}
We first evaluate the frozen vision encoder using behavior cloning (BC). We attach a lightweight MLP policy head to the frozen encoder and train a separate vision-only BC policy for each of the 40 tasks across the four LIBERO suites: Spatial, Object, Goal, and Long. For evaluation, we execute 20 rollouts per task and report the task success rate, corresponding to 200 rollouts per suite.

As shown in Table~\ref{tab:libero_results}, Qwen3VL-8B provides the pretrained VLM encoder reference, while VLM4VLA shows the effect of robot-data pretraining from the same initialization. Our method further improves over vanilla VLM4VLA, suggesting that auxiliary inverse dynamics learning produces visual representations more useful for downstream behavior learning.

\textbf{Behavior cloning loss analysis.}
\begin{table}[t]
  \centering
  \small
  \caption{Frozen-encoder loss analysis on LIBERO. We report average training and validation losses for behavior cloning and proprioceptive state prediction. Lower is better, and bold numbers indicate the lowest loss in each column.}
  \label{tab:frozen_loss_analysis}
  \begin{tabular}{lcccc}
    \toprule
    \multirow{2}{*}{Vision Encoder} 
    & \multicolumn{2}{c}{Behavior Cloning} 
    & \multicolumn{2}{c}{State Prediction} \\
    \cmidrule(lr){2-3} \cmidrule(lr){4-5}
    & Train $\downarrow$ & Val. $\downarrow$ & Train $\downarrow$ & Val. $\downarrow$ \\
    \midrule
    Qwen3VL-8B \citep{bai2025qwen3} & 0.208 & 0.266 & 0.043 & 0.043 \\
    VLM4VLA \citep{zhang2026vlm4vla} & 0.152 & \textbf{0.262} & 0.015 & 0.018 \\
    + Ours & \textbf{0.137} & 0.265 & \textbf{0.009} & \textbf{0.014} \\
    \bottomrule
  \end{tabular}
\end{table}
We further analyze training and validation losses under the same frozen-encoder BC setup. This analysis measures how easily a lightweight policy can fit the action mapping from frozen visual features. If the frozen encoder suffers from state aliasing, visually similar but action-different states are represented by similar features, forcing the BC head to assign different actions to nearly indistinguishable inputs. Therefore, lower training loss indicates that the frozen features provide a less ambiguous action mapping, which is consistent with reduced state aliasing.

As shown in Figure~\ref{fig1:correlation} (b) and Table~\ref{tab:frozen_loss_analysis}, robot-data pretraining reduces training loss compared with the original Qwen3VL-8B encoder, and our method further lowers the average training loss. The validation losses are comparable across the three encoders. These results suggest that inverse dynamics supervision makes the frozen visual features easier to fit for downstream action prediction, providing evidence consistent with reduced state aliasing in the learned representation.

\textbf{Proprioceptive state prediction probe.}
We also probe the frozen vision encoder with proprioceptive state prediction. We train a lightweight MLP regressor on top of frozen visual features to predict robot proprioceptive states using data from all four LIBERO suites, evaluating whether the learned representations preserve robot state information observable from images.

As shown in Figure~\ref{fig1:correlation} (b) and Table~\ref{tab:frozen_loss_analysis}, VLM4VLA reduces state prediction loss compared with the original Qwen3VL-8B encoder, and our method further reduces both training and validation losses. These results indicate that our method better preserves robot state information in the visual representation, providing additional evidence for more precise visual understanding in robotic manipulation.

\subsection{Analysis}
\label{sec4.5:analysis}

\textbf{State-feature alignment.}
\begin{table}[t]
\centering
\caption{State-feature alignment on CALVIN. We report pixel-controlled partial Spearman correlations between visual feature distances and normalized 6-DoF end-effector pose distances using 6,000 frame pairs from CALVIN validation trajectories. Higher is better, and bold numbers indicate the best results.}
\label{tab:state_feature_alignment}
\begin{tabular}{lcc}
\toprule
Model & Mean cosine distance $\uparrow$ & Scale difference $\uparrow$ \\
\midrule
Qwen3VL-8B \citep{bai2025qwen3}  & 0.184 & 0.002 \\
VLM4VLA \citep{zhang2026vlm4vla}  & 0.489 & 0.242 \\
Ours & \textbf{0.600} & \textbf{0.309} \\
\bottomrule
\end{tabular}
\end{table}
We analyze state-feature alignment to examine whether the learned visual representations become more suitable for action prediction. A representation that distinguishes different robot states can reduce state aliasing, where action-relevantly different states are mapped to overly similar features. To measure this, we evaluate whether feature-space distances reflect robot-state distances. Specifically, we sample frame pairs from the same language-annotated trajectory segment and compute a pixel-controlled partial Spearman correlation between the visual feature distance and the normalized 6-DoF end-effector pose distance, while controlling for low-level pixel differences measured by pixel MSE. We report this correlation using two complementary feature metrics: the cosine distance between mean-pooled visual token features and the scale difference between their feature norms. Further details are provided in Appendix~\ref{sup:pixel-controlled-pose-feature-alignment}. As shown in Figure~\ref{fig1:correlation} (c) and Table~\ref{tab:state_feature_alignment}, the pretrained encoder exhibits weak alignment with robot pose changes in both cosine distance and feature scale, suggesting that features learned without robot action training do not sufficiently capture fine-grained robot-state variations. The VLM4VLA baseline improves this alignment, while adding our method further increases the correlation from 0.4886 to 0.5996 for cosine distance and from 0.2424 to 0.3094 for feature-scale difference. These results indicate that our inverse-dynamics supervision makes the visual representation more sensitive to control-relevant state changes beyond raw pixel differences, thereby reducing state aliasing.

\textbf{Ablation study.} We conduct an ablation study on \calvin ~using VLM4VLA to isolate the contribution of Pseudo Time Reversal (PTR), comparing auxiliary inverse dynamics learning (AUX) with and without PTR.
\begin{table}[t]
  \centering
  \small
  \caption{Ablation results on \calvin{} under the same evaluation setting as Table~\ref{tab:calvin_results}. AUX denotes auxiliary inverse dynamics learning without Pseudo Time Reversal, and PTR denotes the full method with Pseudo Time Reversal. We report success rates over 1000 rollouts for completing 1--5 tasks in a row and the average completed sequence length. Bold numbers indicate the best results in each column.}
  \label{tab:calvin_ablation}
  \begin{tabular}{lcccccc}
    \toprule
    Method & 1 & 2 & 3 & 4 & 5 & Avg. Len. $\uparrow$ \\
    \midrule
    VLM4VLA \citep{zhang2026vlm4vla} 
            & 93.4 & 85.9 & 79.7 & 75.2 & 69.2 & 4.03 \\
    + AUX   & 94.3 {\scriptsize (+0.9)}
            & \textbf{87.1} {\scriptsize (+1.2)}
            & 81.1 {\scriptsize (+1.4)}
            & 76.3 {\scriptsize (+1.1)}
            & 70.2 {\scriptsize (+1.0)}
            & 4.09 \\
    + PTR   & \textbf{94.4} {\scriptsize (+0.1)}
            & 86.7 {\scriptsize (-0.4)}
            & \textbf{81.5} {\scriptsize (+0.4)}
            & \textbf{76.6} {\scriptsize (+0.3)}
            & \textbf{70.5} {\scriptsize (+0.3)}
            & \textbf{4.10} \\
    \bottomrule
  \end{tabular}
\end{table}
As shown in Table~\ref{tab:calvin_ablation}, auxiliary inverse dynamics already improves performance over the vanilla baseline, and adding PTR yields a further gain. This result shows that PTR provides a meaningful additional benefit beyond auxiliary inverse dynamics alone.

\section{Limitations \& Future Works}
Our method assumes access to temporally paired observations and actions during training, which is natural for offline robot demonstrations but may be limited in online or streaming settings, even though inverse dynamics supervision can sometimes be constructed from past observations and executed actions. Extending inverse dynamics supervision to such settings remains an important future direction. PTR also relies on pseudo-reversed rather than physically exact reverse actions; future work could explore learned backward dynamics. Finally, PTR currently expands action diversity only through forward and pseudo-reversed directions, and richer transition augmentations may further improve generalization.

\section{Conclusion}

In this paper, we present inverse dynamics learning as an auxiliary objective for VLA vision encoder learning to mitigate state aliasing. By supervising the encoder to predict actions between current and future observations, our method encourages visual representations that preserve fine-grained distinctions needed for low-level control and reduces state aliasing between visually similar states.
We also introduce Pseudo Time Reversal (PTR) to improve generalization under limited action diversity.
After training, the auxiliary head is removed, leaving the main VLA policy architecture and inference process intact. Extensive experiments across robot manipulation benchmarks demonstrate consistent performance gains across diverse VLA baselines. Beyond policy success, frozen-encoder probing and state-feature alignment analysis show that our method learns visual features that are more useful for downstream action and robot state prediction and better aligned with actual robot state changes. These results provide evidence that inverse dynamics supervision reduces state aliasing in the vision encoder, suggesting an effective direction for building more reliable VLA policies.

\bibliographystyle{unsrt}
\bibliography{references}


\appendix

\appendix

\section{Implementation details}
\label{app:implementation_details}

\subsection{Inverse dynamics head}

We implement the inverse dynamics head as a lightweight two-view MLP decoder. Let $Z^{\mathrm{cur}}$ and $Z^{\mathrm{fut}}$ denote the encoded visual token features of the current and future observations, respectively:
\[
Z^{\mathrm{cur}}, Z^{\mathrm{fut}} \in \mathbb{R}^{B \times P \times C},
\]
where \(B\) is the batch size, \(P\) is the number of visual tokens produced by the vision encoder, and \(C\) is the token feature dimension, we first compute their token-wise difference:
\[
Z^{\mathrm{diff}} = Z^{\mathrm{fut}} - Z^{\mathrm{cur}}.
\]
We then concatenate the current, future, and difference features along the channel dimension:
\[
Z^{\mathrm{cat}} = [Z^{\mathrm{cur}}, Z^{\mathrm{fut}}, Z^{\mathrm{diff}}] \in \mathbb{R}^{B \times P \times 3C}.
\]

Each concatenated token is processed independently by a patch-wise fusion MLP, which projects the feature dimension from \(3C\) to a decoder dimension. The resulting fused token features are flattened across the token dimension and passed to an action MLP, which predicts an action chunk:
\[
\bar{a}_{t:t+H} = h_{\psi}(Z^{\mathrm{cur}}, Z^{\mathrm{fut}}) \in \mathbb{R}^{B \times H \times d_a},
\]
where \(H\) denotes the action chunk length chosen for each benchmark and \(d_a=7\) denotes the action dimension.

For observations with multiple input types or camera views, we apply the inverse dynamics head independently to each type. For each input type, we encode the observations at time steps \(t\) and \(t+k\), compute the corresponding inverse dynamics prediction, and evaluate the auxiliary loss using the same action target. The final inverse dynamics loss is averaged across all available input types. The batch size, number of visual tokens, decoder dimension, action chunk length, and other benchmark-specific hyperparameters are reported in the corresponding implementation tables. The inverse dynamics head is used only during training and removed at inference time.

\subsection{Probing MLP heads}

For frozen-encoder evaluation, we attach lightweight MLP probing heads on top of the frozen vision encoder. The vision encoder is kept fixed throughout all probing experiments, and only the newly added MLP head is trained. Both behavior cloning and proprioceptive state prediction use the same probing architecture, differing only in the output dimension and training loss.

Given an input observation, the frozen vision encoder produces visual tokens
\[
Z \in \mathbb{R}^{B \times P \times C},
\]
where \(B\) is the batch size, \(P\) is the number of visual tokens, and \(C\) is the token feature dimension. Each token is first projected independently by a token projection MLP:
\[
Z^{\mathrm{proj}} = \mathrm{GELU}(\mathrm{Linear}(Z)),
\qquad
Z^{\mathrm{proj}} \in \mathbb{R}^{B \times P \times D_{\mathrm{proj}}}.
\]
The projected tokens are then flattened into a single feature vector and passed to a three-layer MLP head:
\[
u = \mathrm{Flatten}(Z^{\mathrm{proj}}),
\]
\[
\hat{y} = f_{\theta}(u).
\]
The MLP head consists of two hidden layers with GELU activations and dropout, followed by a linear output layer:
\[
f_{\theta}: P D_{\mathrm{proj}} \rightarrow 2D_{\mathrm{hidden}} \rightarrow D_{\mathrm{hidden}} \rightarrow d_{\mathrm{out}}.
\]

For the behavior cloning probe, the output is reshaped into an action chunk,
\[
\hat{a}_{t:t+H_{\mathrm{BC}}} \in \mathbb{R}^{B \times H_{\mathrm{BC}} \times d_a},
\]
where \(H_{\mathrm{BC}}\) is the prediction horizon and \(d_a=7\) is the action dimension. For the proprioceptive state prediction probe, the output dimension is set to the target proprioceptive state dimension \(d_s\). In all probing experiments, we use the static RGB view only. We set \(D_{\mathrm{proj}}=256\), \(D_{\mathrm{hidden}}=512\).

\subsection{\calvin{} implementation details}

\calvin{} is a language-conditioned long-horizon manipulation benchmark that evaluates generalization to an unseen environment. Models are trained on demonstrations from environments A, B, and C, and evaluated in environment D. During evaluation, a policy is required to execute a sequence of tabletop manipulation subtasks from visual observations and language instructions. Following the standard protocol, we evaluate each model over 1000 rollouts and report the success rates for completing 1, 2, 3, 4, and 5 consecutive subtasks, as well as the average number of successfully completed subtasks.

For each baseline, we keep the original training pipeline unchanged and add only our auxiliary inverse dynamics loss and PTR. We use the same training budget and evaluation protocol for each baseline and its +Ours variant. Model-specific training details are provided below.

\subsubsection{FLOWER}

\begin{table}[t]
\centering
\caption{Implementation details for FLOWER on CALVIN ABC$\rightarrow$D.}
\label{tab:calvin_flower_impl}
\begin{tabular}{ll}
\toprule
\textbf{Parameter} & \textbf{Value} \\
\midrule
VLM backbone & Florence-2-large \\
Pretrained checkpoint & FLOWER pretrained checkpoint \\
Input views & Static view + wrist view \\
Training steps & 40K \\
Batch size & 32 \\
Optimizer & AdamW \\
Learning rate & $2\times 10^{-5}$ \\
Learning rate schedule & Three-phase schedule $(0.05, 0.1, 0.85)$ \\
Weight decay & 0.05 \\
Action dimension & 7 \\
Action chunk length & 10 \\
Flow sampling steps & 4 \\
DiT dimension / layers / heads & 1024 / 18 / 16 \\
Token dropout & 0.1 \\
Auxiliary loss weight $\lambda_{\mathrm{inv}}$ & 0.3 \\
Gripper loss weight (in auxiliary) $\lambda_g$ & 0.01 \\
PTR probability $p_{\mathrm{rev}}$ & 0.5 \\
Inverse dynamics decoder dimension & 1024 \\
Checkpoint used for evaluation & Final checkpoint \\
\bottomrule
\end{tabular}
\end{table}

Except for the hyperparameters related to our auxiliary objective, we keep the FLOWER training configuration identical to the default hyperparameters released in the official implementation and reported in the original paper. Since our reproduced vanilla FLOWER model did not reach performance comparable to the reported result, we report the publicly available number from the original paper for the vanilla baseline. For our +Ours variant, we train with the same default configuration while adding only auxiliary inverse dynamics learning and PTR. Detailed implementation settings are provided in Table~\ref{tab:calvin_flower_impl}.

The original FLOWER evaluation protocol reports the best checkpoint selected by evaluating every epoch after epoch 19 up to 40K training steps. Since such checkpoint selection can introduce benchmark overfitting, we instead evaluate our +Ours model using the final checkpoint only. Training the +Ours variant takes approximately 4 hours on 4 NVIDIA H100 GPUs.

\subsubsection{VLM4VLA}

\begin{table}[t]
\centering
\caption{Implementation details for VLM4VLA on CALVIN ABC$\rightarrow$D.}
\label{tab:calvin_vlm4vla_impl}
\begin{tabular}{ll}
\toprule
\textbf{Parameter} & \textbf{Value} \\
\midrule
VLM backbone & Qwen3-VL-8B-Instruct \\
Training steps & 30K \\
Batch size & 128 \\
Optimizer & Adam \\
Learning rate & $2\times 10^{-5}$ \\
Weight decay & 0 \\
Warmup epochs & 0.25 \\
Input views & Static view \\
Action dimension & 7 \\
Action chunk length & 10 \\
Auxiliary loss weight $\lambda_{\mathrm{inv}}$ & 0.1 \\
Gripper loss weight (in auxiliary) $\lambda_g$ & 0.01 \\
PTR probability $p_{\mathrm{rev}}$ & 0.5 \\
Inverse dynamics decoder dimension & 1024 \\
Checkpoint used for evaluation & Final checkpoint \\
\bottomrule
\end{tabular}
\end{table}

For VLM4VLA on \calvin{}, we follow the released configuration as closely as possible and keep the same hyperparameters for the vanilla baseline and its +Ours variant, except for the auxiliary inverse dynamics objective and PTR. The model is built on Qwen3-VL-8B-Instruct and uses the static camera view as input. Following the original paper, we evaluate both models at the 30K checkpoint. Detailed implementation settings are provided in Table~\ref{tab:calvin_vlm4vla_impl}. Training each VLM4VLA model on CALVIN takes approximately 20 hours on 8 NVIDIA H100 GPUs.

\subsection{Bridge / SimplerEnv implementation details}

SimplerEnv is a real-to-sim benchmark that evaluates policies trained on real-world robot data in simulated environments. It provides evaluation settings based on different data sources, including Google Robot and WidowX Bridge. Following prior VLA evaluations, we focus on the WidowX Bridge setting, which is based on the Bridge dataset.

\begin{table}[t]
\centering
\caption{Implementation details for SpatialVLA on Bridge / SimplerEnv.}
\label{tab:bridge_spatialvla_impl}
\begin{tabular}{ll}
\toprule
\textbf{Parameter} & \textbf{Value} \\
\midrule
VLM backbone & PaliGemma2 \\
Pretrained checkpoint & SpatialVLA pretrained checkpoint \\
Training steps & 30K \\
Batch size & 256 \\
Optimizer & AdamW \\
Learning rate & $2\times 10^{-5}$ \\
Weight decay & 0 \\
Warmup ratio & 0.005 \\
Input views & Static view \\
Action dimension & 7 \\
Action chunk length & 4 \\
Auxiliary loss weight $\lambda_{\mathrm{inv}}$ & 0.2 \\
Gripper loss weight (in auxiliary) $\lambda_g$ & 0.01 \\
PTR probability $p_{\mathrm{rev}}$ & 0.5 \\
Inverse dynamics decoder dimension & 512 \\
Checkpoint used for evaluation & Final checkpoint \\
Execution steps & 1 \\
\bottomrule
\end{tabular}
\end{table}

We use SpatialVLA as the main baseline for SimplerEnv. We initialize from the released SpatialVLA pretrained checkpoint, finetune the model on Bridge, and evaluate it on four WidowX manipulation tasks: putting a spoon on a towel, putting a carrot on a plate, stacking a green block on a yellow block, and putting an eggplant into a yellow basket. For each task, we run 24 trials with randomized initializations and report the final task success rate. Detailed implementation settings are provided in Table~\ref{tab:bridge_spatialvla_impl}.

Except for the auxiliary inverse dynamics objective and PTR used in our +Ours variant, we keep all hyperparameters identical to the official code. Since the exact training steps and checkpoint selection protocol are not specified in the official release, we train SpatialVLA for 30K steps, matching the VLM4VLA training budget used in our CALVIN experiments, and evaluate the final checkpoint. For comparison, the OpenVLA and RoboVLMs results in Table~\ref{tab:simplerenv_bridge} are taken from \citep{qu2025spatialvla}. We also re-evaluate the publicly released SpatialVLA pretrained checkpoint under the same SimplerEnv evaluation protocol. Training each SpatialVLA finetuning run takes approximately 16 hours on 4 NVIDIA B200 GPUs.

\subsection{LIBERO-90 pretraining}

\begin{table}[t]
\centering
\caption{Implementation details for LIBERO-90 pretraining.}
\label{tab:libero90_pretrain_impl}
\begin{tabular}{ll}
\toprule
\textbf{Parameter} & \textbf{Value} \\
\midrule
VLM backbone & Qwen3-VL-8B-Instruct \\
Training epochs & 1 \\
Batch size & 256 \\
Optimizer & Adam \\
Learning rate & $2\times 10^{-5}$ \\
Weight decay & 0 \\
Warmup epochs & 0.25 \\
Input views & Static view \\
Action dimension & 7 \\
Action chunk length & 4 \\
Auxiliary loss weight $\lambda_{\mathrm{inv}}$ & 0.1 \\
Gripper loss weight $\lambda_g$ & 0.01 \\
PTR probability $p_{\mathrm{rev}}$ & 0.5 \\
Inverse dynamics decoder dimension & 1024 \\
Checkpoint used for evaluation & Final checkpoint \\
\bottomrule
\end{tabular}
\end{table}

For LIBERO-90 pretraining, we use VLM4VLA with Qwen3-VL-8B-Instruct as the backbone and train on LIBERO-90 for one epoch. Except for the auxiliary inverse dynamics objective and PTR, we keep the same hyperparameters for the vanilla baseline and its +Ours variant. Both models are trained with the static camera view and evaluated using the final checkpoint for the frozen-encoder analyses. Detailed implementation settings are provided in Table~\ref{tab:libero90_pretrain_impl}. Training each LIBERO-90 pretraining model takes approximately 12 hours on 8 NVIDIA H100 GPUs.

\subsection{Behavior cloning probe}

\begin{table}[t]
\centering
\caption{Implementation details for the behavior cloning probe.}
\label{tab:bc_probe_impl}
\begin{tabular}{ll}
\toprule
\textbf{Parameter} & \textbf{Value} \\
\midrule
Input views & Static view \\
Token projection dimension & 256 \\
MLP hidden dimension & 512 \\
Action dimension & 7 \\
Action chunk length & 4 \\
Output shape & $B \times (4 \times 7$) \\
Loss & $\ell_1$ + BCE \\
Gripper loss weight $\lambda_g$ & 0.01 \\
Training loss reporting & Average over 1K steps \\
Validation loss reporting & Average of validation losses every 100 steps \\
Optimizer & Adam \\
Learning rate & $2\times10^{-4}$ \\
Batch size & 64 \\
Training steps & 1K \\
Evaluation tasks & 40 tasks across LIBERO-Spatial, Object, Goal, and Long \\
Evaluation rollouts & 20 per task \\
Checkpoint used for evaluation & Final checkpoint \\
\bottomrule
\end{tabular}
\end{table}

For the behavior cloning probe, we freeze the vision encoder pretrained on LIBERO-90 and train only a lightweight MLP policy head. The probe uses visual features from the static RGB view and predicts a 4-step action chunk with 7-dimensional actions. The policy head is trained with an $\ell_1$ loss on the motion offsets and a binary cross-entropy loss on the gripper action, with the gripper loss weighted by $\lambda_g=0.01$. Detailed implementation settings are provided in Table~\ref{tab:bc_probe_impl}.

We train a separate behavior cloning probe for each of the 40 tasks across LIBERO-Spatial, Object, Goal, and Long. For each task, we use 10\% of the demonstrations as the validation split. Each probe is trained for 1K steps with Adam using a learning rate of $2\times10^{-4}$ and batch size 64. For the loss analysis, we report the average training loss over the full 1K training steps and the average validation loss computed from evaluations performed every 100 steps. For rollout evaluation, we use the final checkpoint and execute 20 rollouts per task. Training each behavior cloning probe takes approximately 1 hour on a single NVIDIA RTX 3090 GPU.

\subsection{Proprioceptive state prediction probe}

\begin{table}[t]
\centering
\caption{Implementation details for the proprioceptive state prediction probe.}
\label{tab:state_probe_impl}
\begin{tabular}{ll}
\toprule
\textbf{Parameter} & \textbf{Value} \\
\midrule
Input views & Static view \\
Token projection dimension & 256 \\
MLP hidden dimension & 512 \\
Target dimension & 8 \\
Target state & 3D EEF position + 3D axis-angle orientation + 2D gripper qpos \\
Loss & $\ell_1$ \\
Optimizer & Adam \\
Learning rate & $2\times10^{-4}$ \\
Batch size & 256 \\
Training epochs & 1 \\
Training loss reporting & Recorded at every training step \\
Validation loss reporting & Evaluated every 100 steps \\
\bottomrule
\end{tabular}
\end{table}

For the proprioceptive state prediction probe, we freeze the vision encoder pretrained on LIBERO-90 and train only a lightweight MLP regression head. The probe uses visual features from the static RGB view and predicts an 8-dimensional proprioceptive state. The target state consists of the 3D end-effector position, 3D axis-angle orientation, and 2D gripper qpos. We train the probe with an $\ell_1$ regression loss. Detailed implementation settings are provided in Table~\ref{tab:state_probe_impl}.

This probe evaluates whether the frozen visual representations preserve robot state information that is observable from the image. Lower prediction loss indicates that the learned visual features encode more precise information about the robot configuration, complementing the behavior cloning evaluation. Training the proprioceptive state prediction probe takes approximately 3 hours on a single NVIDIA RTX 3090 GPU.

\subsection{State-feature alignment analysis}
\label{sup:pixel-controlled-pose-feature-alignment}

We evaluate whether visual feature distances preserve robot pose geometry after
accounting for low-level image differences. The analysis uses language-annotated
CALVIN validation trajectories. We first collect all frames that belong to
language annotation intervals, resulting in 64,793 annotated validation records.
Frame pairs are sampled only within the same language-annotated trajectory
segment, so the two frames share the same instruction and high-level task
context while the robot state varies. We use temporal gaps
$\{1,2,4,8,16,32\}$ and uniformly sample 1,000 pairs for each gap, yielding
6,000 frame pairs in total.

For each frame $i$, we use the 6-DoF end-effector pose $s_i$, corresponding to
the first six dimensions of \texttt{robot\_obs}. Let
$\sigma \in \mathbb{R}^6$ denote the per-dimension standard deviation of these
six pose dimensions over all annotated validation records. The target pose
distance for a pair $(i,j)$ is
\begin{equation}
  d_{\mathrm{pose}}(i,j)
  =
  \left\|
  \frac{s_i - s_j}{\sigma}
  \right\|_2.
  \label{eq:app-pose6-z-l2}
\end{equation}

For visual features, we extract spatial visual tokens from the selected vision
encoder layer. The main analysis uses layer 24. We evaluate two types of
feature distance: cosine distance, which measures directional feature change,
and scale distance, which measures change in feature magnitude. Let $h_i^k$
and $h_j^k$ be the visual tokens at spatial position $k$, and let
$\bar{h}_i=\frac{1}{K}\sum_k h_i^k$ be the mean-pooled visual feature. The
pooled cosine and pooled scale distances are defined as
\begin{align}
  d_{\mathrm{cos}}(i,j)
  &=
  1 -
  \frac{\langle \bar{h}_i, \bar{h}_j \rangle}
       {\|\bar{h}_i\|_2 \|\bar{h}_j\|_2},
  \label{eq:app-pooled-cosine-distance}
  \\
  d_{\mathrm{scale}}(i,j)
  &=
  \left| \|\bar{h}_i\|_2 - \|\bar{h}_j\|_2 \right|.
  \label{eq:app-pooled-scale-distance}
\end{align}

To remove the effect of low-level image change, we compute a pixel-controlled
partial Spearman correlation. For each pair, $d_{\mathrm{pix}}$ is computed as
\texttt{pixel\_mse} from downsampled grayscale image thumbnails. We
rank-transform feature distance, pose distance, and $d_{\mathrm{pix}}$.
Ranked feature distance and ranked pose distance are then separately
residualized against ranked $d_{\mathrm{pix}}$, and the final score is the
Pearson correlation between the two residual vectors:
\begin{equation}
  \rho_{\mathrm{partial}}
  =
  \mathrm{corr}
  \left(
  \mathrm{resid}(\mathrm{rank}(d_{\mathrm{feat}}), \mathrm{rank}(d_{\mathrm{pix}})),
  \mathrm{resid}(\mathrm{rank}(d_{\mathrm{pose}}), \mathrm{rank}(d_{\mathrm{pix}}))
  \right).
  \label{eq:app-partial-spearman}
\end{equation}
This score measures whether feature distance tracks 6-DoF end-effector pose
change beyond what can be explained by raw pixel-level image difference.

\section{Qualitative results}

\begin{figure}[t]
        \centering
        \includegraphics[width=\linewidth]{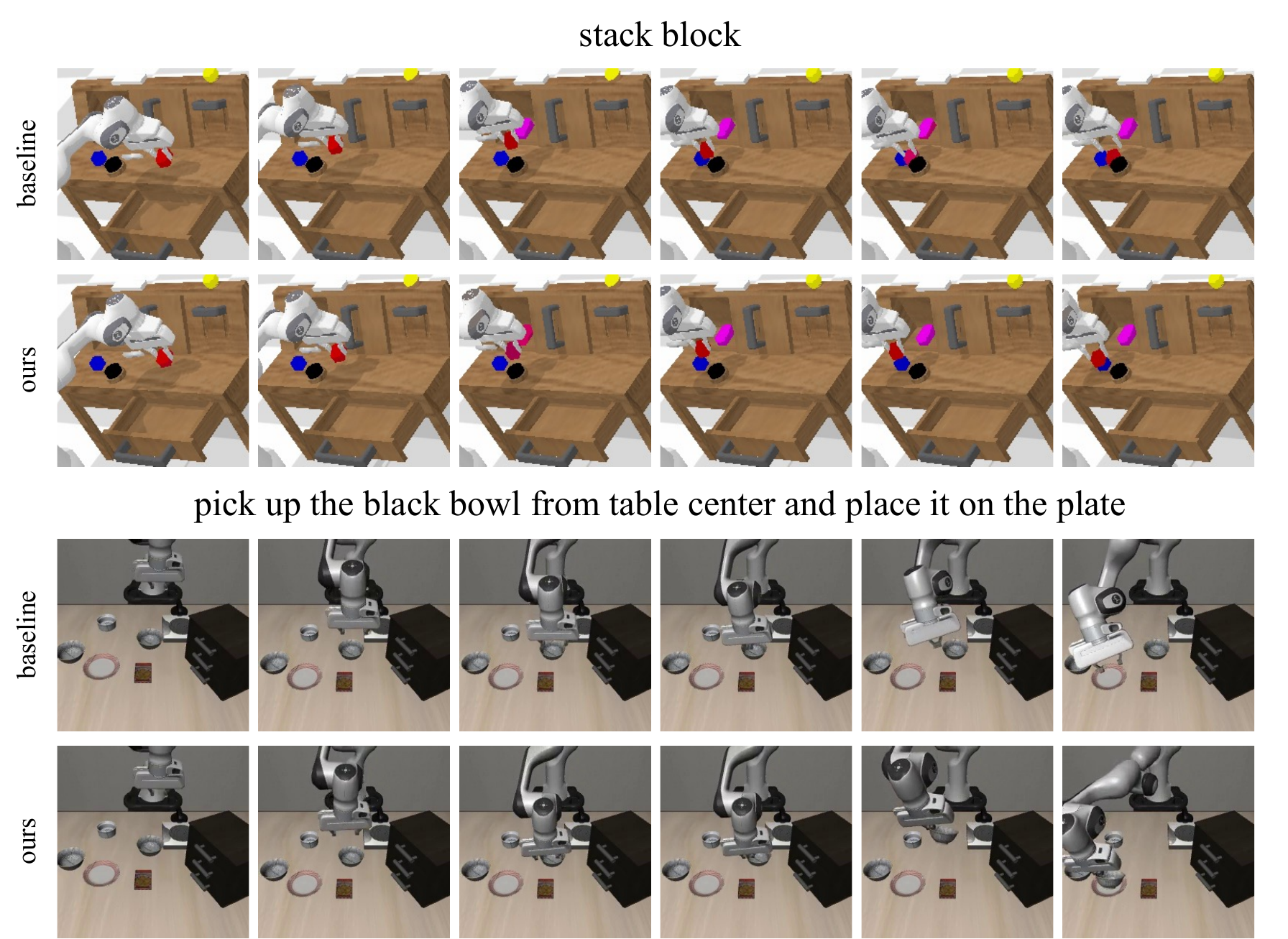}
        \caption{
        Qualitative results on CALVIN ABC$\rightarrow$D and LIBERO. 
        Each rollout is shown with six representative frames. 
        \textbf{Top:} In the CALVIN stack-block task, the vanilla VLM4VLA policy releases the red block before it is properly aligned above the blue block, while our method accurately places the red block on the target. 
        \textbf{Bottom:} In the LIBERO-Object task, the vanilla frozen-encoder BC policy approaches the black bowl imprecisely and fails to grasp it, whereas our method moves to the correct grasp region, picks up the bowl, and places it on the plate.
        }
        \label{fig:qualitative_results}
\end{figure}

Figure~\ref{fig:qualitative_results} shows qualitative examples from CALVIN ABC$\rightarrow$D and LIBERO. Each row visualizes six representative frames from a rollout. The examples illustrate that our method leads to more precise behavior compared to the vanilla VLM4VLA baseline.

In the CALVIN stack-block task, the vanilla VLM4VLA policy moves the red block toward the blue block but lowers the end-effector and releases the block before it is sufficiently aligned with the target, resulting in failure. In contrast, our method moves the red block accurately above the blue block and successfully places it on the target block.

In the LIBERO-Object task, we evaluate the frozen-encoder BC probe on the instruction "pick up the black bowl from table center and place it on the plate." The vanilla encoder-based policy moves only coarsely toward the black bowl and lowers the gripper without accurately reaching the grasp point, causing the policy to miss the object before moving toward the plate. In contrast, the policy trained on our encoder moves directly to the correct grasp region, successfully picks up the bowl, and transfers it to the target plate.

% ...

\section{Broader impacts}

Our work aims to improve visual representation learning for vision-language-action (VLA) models in robot manipulation. A potential positive impact is that stronger and more reliable visual representations may improve the robustness and generalization of robot policies, which could benefit assistive robotics, household automation, and industrial manipulation systems. Because our method is plug-and-play and uses existing observation-action data, it may also lower the cost of improving VLA models without requiring additional annotations.

At the same time, improved robot manipulation policies may raise safety and misuse concerns if deployed without appropriate safeguards. More capable VLA policies could be used in settings where robots interact with people, fragile objects, or safety-critical environments, and failures caused by distribution shift, incorrect language instructions, or visual ambiguity could lead to unintended physical actions. The method could also improve manipulation capabilities in applications that require careful access control. We therefore emphasize that deployment should include task-specific safety constraints, human oversight, controlled evaluation under distribution shifts, and appropriate restrictions on use in safety-critical or potentially harmful settings.

\section{Licenses for existing assets}
\begin{table}[t]
\centering
\caption{Existing assets used in this work.}
\label{tab:asset_licenses}
\small
\begin{tabularx}{\linewidth}{p{0.18\linewidth} p{0.18\linewidth} p{0.28\linewidth} X}
\toprule
Asset & Type & Usage in this work & License / terms \\
\midrule
CALVIN & Benchmark / simulator / Dataset & CALVIN ABC$\rightarrow$D evaluation & MIT License, Copyright (c) 2021 Oier Mees \\
SimplerEnv & Benchmark / simulator & WidowX simulation evaluation & MIT License, Copyright (c) 2024 simpler-env \\
BridgeData V2 / Bridge & Dataset & Training data for SimplerEnv baselines & Creative Commons Attribution 4.0 International (CC BY 4.0) \\
LIBERO / LIBERO-90 & Benchmark / dataset & Frozen-encoder evaluation and pretraining & MIT License, Copyright (c) 2023 Lifelong Robot Learning \\
Qwen3-VL-8B-Instruct & Pretrained model & VLM4VLA backbone for CALVIN and LIBERO & Apache License 2.0 \\
Qwen3-VL-4B-Instruct & Pretrained model & VLM4VLA backbone for SimplerEnv & Apache License 2.0 \\
VLM4VLA & Baseline implementation & CALVIN, and LIBERO experiments & Not explicitly specified in the official repository \\
FLOWER & Baseline implementation / Pretrained model & CALVIN baseline and +Ours experiments & MIT License, Copyright (c) 2025 Moritz Reuss \\
SpatialVLA & Baseline implementation / Pretrained model & SimplerEnv baseline and +Ours experiments & MIT License; third-party code/models subject to respective licenses \\
\bottomrule
\end{tabularx}
\end{table}

We use publicly available datasets, benchmarks, pretrained models, and baseline implementations only for research purposes, following their respective licenses and terms of use. We cite the original papers and repositories for all assets used in our experiments. Table~\ref{tab:asset_licenses} summarizes the main assets used in this work.



\end{document}